\documentclass{article}

\PassOptionsToPackage{numbers,sort&compress}{natbib}
\usepackage[preprint]{nips_2018}

\usepackage{amsmath}
\usepackage{graphicx}
\usepackage{bm}
\usepackage{amsfonts}
\usepackage{mathtools}
\usepackage{adjustbox}
\usepackage{tabularx}
\usepackage{multirow}
\usepackage{makecell}
\usepackage{hhline}
\usepackage{tikz}
\usepackage{hyperref}
\usepackage{colortbl}

\graphicspath{ {images/} }

\usepackage{alphalph}
\usepackage{caption}
\usepackage{subcaption}

\usetikzlibrary{arrows.meta}
\usetikzlibrary{positioning}
\setcitestyle{square}

\def\x{{\mathbf x}}
\def\y{{\mathbf y}}
\def\m{{\mathbf m}}
\def\h{{\mathbf h}}
\def\W{{\mathbf W}}
\def\U{{\mathbf U}}
\def\M{{\mathbf M}}
\def\E{{\text E}}
\def\Var{{\text{Var}}}
\def\ident{{\mathbf I}}

\begin{document}

\title{Survey of Dropout Methods for Deep Neural Networks}

\author{Alex Labach \\ University of Toronto \\ \texttt{alex.labach@mail.utoronto.ca} \\ \And Hojjat Salehinejad \\ University of Toronto \\ \texttt{hojjat.salehinejad@mail.utoronto.ca} \\ \And Shahrokh Valaee \\ University of Toronto \\ \texttt{valaee@ece.utoronto.ca}}
\maketitle

\begin{abstract}

Dropout methods are a family of stochastic techniques used in neural network training or inference that have generated significant research interest and are widely used in practice. They have been successfully applied in various applications, including neural network regularization, model compression, and in measuring the uncertainty of neural network outputs. While originally formulated for dense neural network layers, recent advances have made dropout methods also applicable to convolutional and recurrent neural network layers. This paper summarizes the history of dropout methods, their various applications, and current areas of research interest. Important proposed methods are described in additional detail.

\end{abstract}

\section{Introduction}

Deep neural networks are a topic of widespread interest in contemporary artificial intelligence and signal processing. Their high number of parameters make them particularly prone to overfitting, requiring regularization methods in practice. Dropout was introduced in 2012 as a technique to avoid overfitting~\cite{hinton2012improving} and was subsequently applied in the 2012 winning submission for the Large Scale Visual Recognition Challenge that revitalized deep neural network research~\cite{krizhevsky2012imagenet}. The original method omitted each neuron in a neural network with probability 0.5 during each training iteration, with all neurons being included during testing. This technique was shown to significantly improve results on a variety of tasks~\cite{hinton2012improving}.

\begin{figure}[!t]
\centering
\captionsetup{font=small}
\includegraphics[width=1\textwidth]{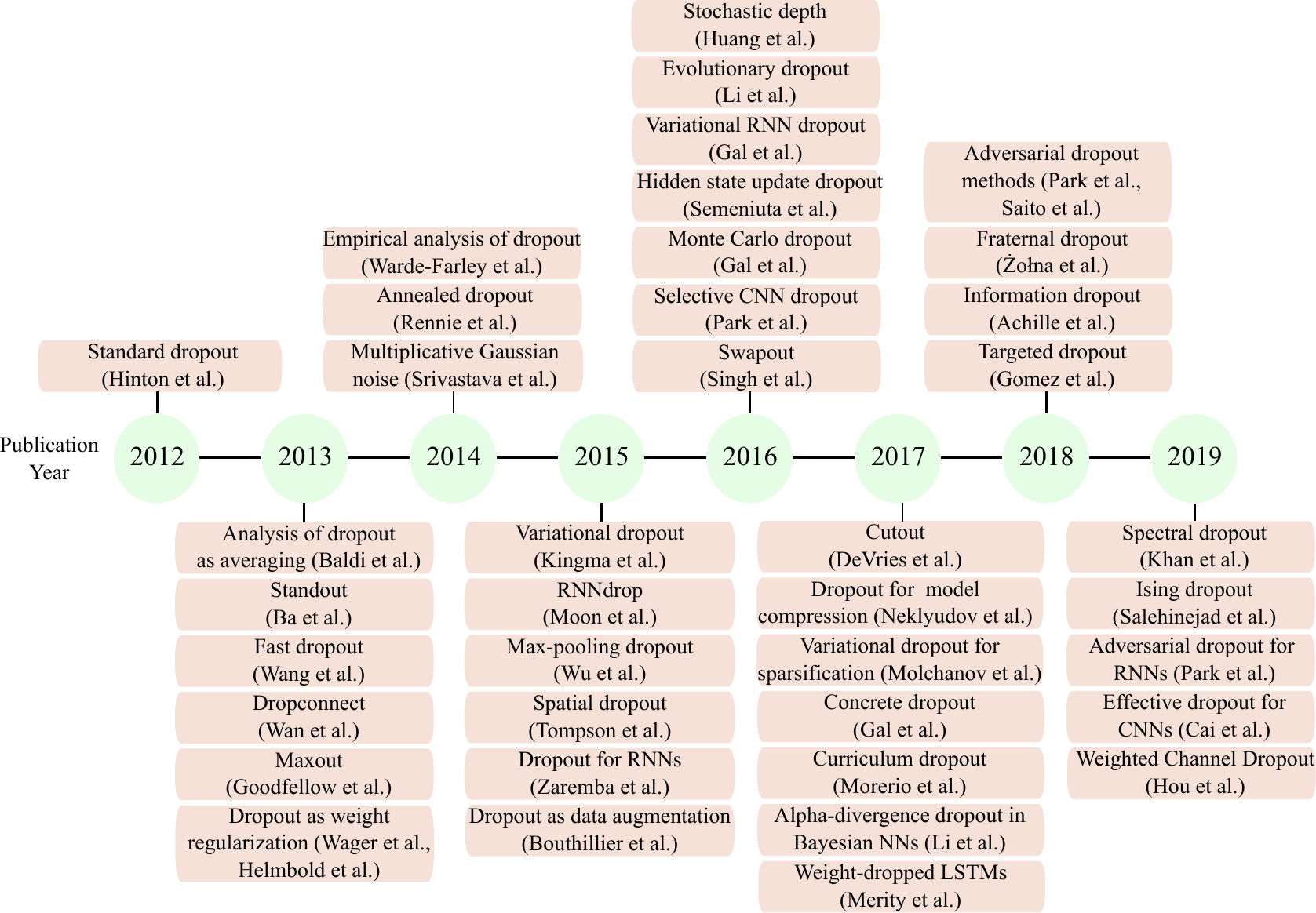}   
\caption{Some proposed methods and theoretical advances in dropout methods from 2012 to 2019.}
\label{fig:advances}
\end{figure}

In the years since, a wide range of stochastic techniques inspired by the original dropout method have been proposed for use with deep learning models. We use the term \textit{dropout methods} to refer to them in general. They include dropconnect~\cite{wan2013dropconnect}, standout~\cite{ba2013adaptive}, fast dropout~\cite{wang2013fast}, variational dropout~\cite{kingma2015variational}, Monte Carlo dropout~\cite{gal2016dropout} and many others. Generally speaking, dropout methods involve randomly modifying neural network parameters or activations during training or inference, or approximating this process. Figure \ref{fig:advances} illustrates research into dropout methods over time.

While originally used to avoid overfitting, dropout methods have since expanded to a variety of applications. The two additional applications discussed in this paper are the use of dropout to compress deep neural networks~\cite{molchanov2017variational,neklyudov2017structured,salehinejad2019ising,gomez2018targeted} and Monte Carlo dropout~\cite{gal2016dropout}, which measures the uncertainty of deep learning models during inference.

Another direction of research into dropout methods has been applying them to a wider range of neural network topologies. This includes methods for applying dropout to convolutional neural network layers~\cite{wu2015towards,tompson2015efficient,park2016analysis,devries2017improved,huang2016deep,cai2019effective,khan2019regularization,hou2019weighted} as well as to recurrent neural networks (RNNs)~\cite{moon2015rnndrop,gal2016rnn,semeniuta2016recurrent,krueger2016zoneout,merity2017regularizing}. RNN dropout methods in particular have become commonly used, and have been recently applied in improving state-of-the-art results in natural language processing~\cite{merity2017regularizing, melis2017neural,zolna2018fraternal}.

This paper provides an overview of past and current research into dropout methods. Although not exhaustive, certain dropout methods that are particularly influential or representative of particular areas of research are described in detail. The rest of the paper is structured as follows. Section \ref{sec:standard} describes the original dropout method proposed by \citet{hinton2012improving} and introduces basic concepts common to dropout methods. Section \ref{sec:theoretical} summarizes approaches for theoretically explaining the function of dropout methods. Section \ref{sec:training} describes dropout methods for general neural network training. Section \ref{sec:cnn} describes dropout methods specialized for training convolutional neural network layers and Section \ref{sec:rnn} describes methods specialized for recurrent neural network layers. Section \ref{sec:compression} summarizes dropout methods for compressing neural networks. Section \ref{sec:mc} describes Monte Carlo dropout and related work. Finally, Section \ref{sec:discussion} discusses current and future research directions.

\subsection{Notation}

We follow a number of common conventions when providing formulas describing neural network cells or layers. Bold lower-case letters represent vectors and bold upper-case letters represent matrices. Most of the time, when multiplying an input vector by a weight matrix, a vector of learned biases may also be added, for example producing \(\W\x+\mathbf{b}\) from an input vector \(\x\). To simplify notation, we generally treat the biases as elements of the weight matrix, with an element with value 1 implicitly appended to the vector \(\x\). So, the previous operation would be written as \(\W\x\). The operator \(\circ\) represents element-wise (or Hadamard) multiplication.

\section{Standard dropout} \label{sec:standard}

\begin{figure}[!t]
	\centering
	\begin{tikzpicture}[]
		\node[circle,minimum size=15,draw](1) at (0,0) {};
		\node[circle,minimum size=15,draw](2) at (0,0.75) {};
		\node[circle,minimum size=15,draw](3) at (0,1.5) {};
		\node[circle,minimum size=15,draw](4) at (0,2.25) {};
		\node[circle,minimum size=15,draw](5) at (0,3) {};
		\node[circle,minimum size=15,draw](6) at (1.5,0) {};
		\node[circle,minimum size=15,draw](7) at (1.5,0.75) {};
		\node[circle,minimum size=15,draw](8) at (1.5,1.5) {};
		\node[circle,minimum size=15,draw](9) at (1.5,2.25) {};
		\node[circle,minimum size=15,draw](10) at (1.5,3) {};
		\node[circle,minimum size=15,draw](11) at (3,1) {};
		\node[circle,minimum size=15,draw](12) at (3,2) {};
		\draw [-Stealth] (1) -- (6); \draw [-Stealth] (1) -- (7); \draw [-Stealth] (1) -- (8); \draw [-Stealth] (1) -- (9); \draw [-Stealth] (1) -- (10);
		\draw [-Stealth] (2) -- (6); \draw [-Stealth] (2) -- (7); \draw [-Stealth] (2) -- (8); \draw [-Stealth] (2) -- (9); \draw [-Stealth] (2) -- (10);
		\draw [-Stealth] (3) -- (6); \draw [-Stealth] (3) -- (7); \draw [-Stealth] (3) -- (8); \draw [-Stealth] (3) -- (9); \draw [-Stealth] (3) -- (10);
		\draw [-Stealth] (4) -- (6); \draw [-Stealth] (4) -- (7); \draw [-Stealth] (4) -- (8); \draw [-Stealth] (4) -- (9); \draw [-Stealth] (4) -- (10);
		\draw [-Stealth] (5) -- (6); \draw [-Stealth] (5) -- (7); \draw [-Stealth] (5) -- (8); \draw [-Stealth] (5) -- (9); \draw [-Stealth] (5) -- (10);
		\draw [-Stealth] (6) -- (11); \draw [-Stealth] (6) -- (12);
		\draw [-Stealth] (7) -- (11); \draw [-Stealth] (7) -- (12);
		\draw [-Stealth] (8) -- (11); \draw [-Stealth] (8) -- (12);
		\draw [-Stealth] (9) -- (11); \draw [-Stealth] (9) -- (12);
		\draw [-Stealth] (10) -- (11); \draw [-Stealth] (10) -- (12);
	\end{tikzpicture}
	\hspace{5mm}
	\begin{tikzpicture}[]
		\node[circle,minimum size=15,draw](1) at (0,0) {};
		\node[circle,minimum size=15,draw](2) at (0,0.75) {};
		\node[circle,minimum size=15,fill=black,draw](3) at (0,1.5) {};
		\node[circle,minimum size=15,draw](4) at (0,2.25) {};
		\node[circle,minimum size=15,fill=black,draw](5) at (0,3) {};
		\node[circle,minimum size=15,fill=black,draw](6) at (1.5,0) {};
		\node[circle,minimum size=15,draw](7) at (1.5,0.75) {};
		\node[circle,minimum size=15,fill=black,draw](8) at (1.5,1.5) {};
		\node[circle,minimum size=15,fill=black,draw](9) at (1.5,2.25) {};
		\node[circle,minimum size=15,draw](10) at (1.5,3) {};
		\node[circle,minimum size=15,draw](11) at (3,1) {};
		\node[circle,minimum size=15,draw](12) at (3,2) {};
		\draw [-Stealth] (1) -- (7); \draw [-Stealth] (1) -- (10);
		\draw [-Stealth] (2) -- (7); \draw [-Stealth] (2) -- (10);
		\draw [-Stealth] (4) -- (7); \draw [-Stealth] (4) -- (10);
		\draw [-Stealth] (7) -- (11); \draw [-Stealth] (7) -- (12);
		\draw [-Stealth] (10) -- (11); \draw [-Stealth] (10) -- (12);
	\end{tikzpicture}
	\caption{An example of standard dropout. The left network is fully connected, and the right has had neurons dropped with probability 0.5. Dropout is not applied to the output layer.}
	\label{fig:standard_dropout}
\end{figure}
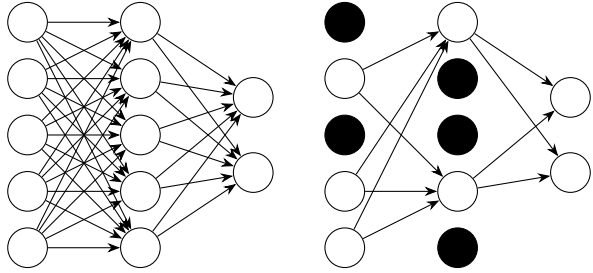

The original proposed dropout method, introduced by \citet{hinton2012improving} in 2012, provides a simple technique for avoiding overfitting in feedforward neural networks. During each training iteration, each neuron is omitted from the network with probability \(p\). Once trained, the full network is used, although neuron outputs are multiplied by the probability \(p\) that the neuron was omitted. This compensates for the larger size of the network now that no neurons are dropped, and can be interpreted as averaging over the possible networks during training. The probability can vary for each layer, with the original paper recommending \(p=0.2\) for the input layer and \(p=0.5\) for hidden layers. Neurons in the output layer are not dropped. This technique is usually simply known as dropout, but for the purposes of this article we will call it \textit{standard dropout}, to distinguish it from other dropout methods. This method is illustrated in Figure~\ref{fig:standard_dropout}.

Mathematically, the behaviour of standard dropout during training for a neural network layer is given by:
\begin{equation} 
\label{eq:standard}
	\y = f(\W\x) \circ \m, \quad m_i \sim \text{Bernoulli}(1-p)
\end{equation}
where \(\y\) is the layer output, \(f(\cdot)\) is the activation function, \(\W\) is the layer weight matrix, \(\x\) is the layer input, and \(\m\) is the layer dropout mask, with each element \(m_i\) being 0 with probability \(p\). Once trained, the layer output is given by
\begin{equation}
	\y = (1-p)f(\W\x).
\end{equation}

Standard dropout is equivalent to adding an additional layer after a layer of neurons that simply sets values to zero with some probability during training, and multiplies them by \(1-p\) during testing. Other formulations of standard dropout may scale weights rather than outputs during testing, or scale outputs by \(1/(1-p)\) during training rather than scaling them during testing, but both of these approaches have the same effect as the formulation given here.

This method proved effective for regularizing neural networks, enabling them to be trained for longer periods without overfitting and resulting in improved test accuracy~\cite{hinton2012improving,srivastava2014dropout}. Standard dropout has since become widely used in practice.



\section{Theoretical understandings of dropout} \label{sec:theoretical}

Substantial theoretical work has been done to understand why standard dropout works, how it affects neural network training, and to establish links with other concepts and techniques in deep learning. Two important directions in this area have been interpreting dropout as implicitly averaging over an ensemble of neural networks, and linking neural networks with dropout to Bayesian machine learning models.

In the original dropout paper, \citet{hinton2012improving} observe that there is a large number of possible neural network structures that result from randomly dropping neurons, and suggest that dropout implicitly performs averaging over this ensemble of possible networks. For instance, using a network with a single hidden layer of \(N\) units and softmax activation is equivalent to taking the geometric mean of the outputs of the \(2^N\) possible networks under dropout~\cite{hinton2012improving}. This is similar to bagging, a machine learning technique where multiple instances of a model are trained separately and the arithmetic average of their output is used in inference. Standard dropout training varies from bagging in that only one model is trained, and an approximation to the geometric mean of the dropout ensemble's outputs is used rather than an arithmetic mean~\cite{wardefarley2014empirical}. Later work has built a firmer foundation for this interpretation by analysing the suitability of the geometric mean and the quality of the approximation, both mathematically and empirically~\cite{baldi2013understanding,wardefarley2014empirical}.

Another theoretical approach links dropout methods to Bayesian machine learning. An ideal Bayesian model places a prior distribution over the model parameters, then determines the posterior distribution of parameters given a training set, and marginalizes over this distribution to perform inference on an input. In practice, this is computationally expensive, and so approximations are used to simplify this process. Various authors have argued that training with dropout methods can be interpreted as using a Bayesian model with certain approximations~\cite{ba2013adaptive,wang2013fast,gal2016dropout}. This provides a justification for using dropout methods grounded in probability theory. The Bayesian interpretation of dropout presented by \citet{gal2016dropout} has been particularly influential. The authors show that training a neural network with standard dropout is equivalent to optimizing a variational objective between an approximate distribution and the posterior of a deep Gaussian process, which is a Bayesian machine learning model. This insight led to the development of Monte Carlo dropout, described in Section~\ref{sec:mc}.

Although the two approaches described above have been widely applied, neither has completely dominated research into dropout methods, and various alternative approaches that seek to link dropout to established machine learning techniques or concepts have been explored. These include analyzing dropout as a weight regularization method~\cite{wager2013dropout,helmbold2015inductive}, as a data augmentation method~\cite{bouthillier2015dropout}, and in terms of information theory~\cite{achille2018information}.

Research into standard dropout has also shown empirical properties that have proven useful in understanding dropout and in developing new dropout methods. \citet{hinton2012improving} showed that standard dropout reduces feature co-adaptation, where the outputs of individual neurons only provide useful information in combination with other neuron outputs. They argue that reducing co-adaptation leads to improved generalization. \citet{srivastava2014dropout} showed that standard dropout also promotes sparsity in the weights of neural networks, causing more weights to be near zero. This has led to research interest in using dropout to sparsify and ultimately compress neural networks, which is described in Section \ref{sec:compression}.

\section{Dropout methods for training} \label{sec:training}

This section describes significant dropout methods that, like standard dropout, regularize dense feedforward neural network layers during training. Most of these methods were directly inspired by standard dropout, and seek to improve on its speed or regularization effectiveness. Dropout methods for other kinds of neural network layers or for applications other than regularization are described in later sections.

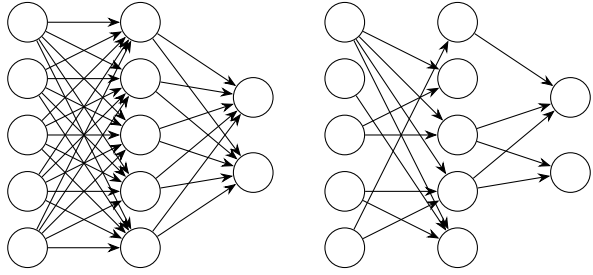
\begin{figure}[!t]
	\centering
	\begin{tikzpicture}[]
		\node[circle,minimum size=15,draw](1) at (0,0) {};
		\node[circle,minimum size=15,draw](2) at (0,0.75) {};
		\node[circle,minimum size=15,draw](3) at (0,1.5) {};
		\node[circle,minimum size=15,draw](4) at (0,2.25) {};
		\node[circle,minimum size=15,draw](5) at (0,3) {};
		\node[circle,minimum size=15,draw](6) at (1.5,0) {};
		\node[circle,minimum size=15,draw](7) at (1.5,0.75) {};
		\node[circle,minimum size=15,draw](8) at (1.5,1.5) {};
		\node[circle,minimum size=15,draw](9) at (1.5,2.25) {};
		\node[circle,minimum size=15,draw](10) at (1.5,3) {};
		\node[circle,minimum size=15,draw](11) at (3,1) {};
		\node[circle,minimum size=15,draw](12) at (3,2) {};
		\draw [-Stealth] (1) -- (6); \draw [-Stealth] (1) -- (7); \draw [-Stealth] (1) -- (8); \draw [-Stealth] (1) -- (9); \draw [-Stealth] (1) -- (10);
		\draw [-Stealth] (2) -- (6); \draw [-Stealth] (2) -- (7); \draw [-Stealth] (2) -- (8); \draw [-Stealth] (2) -- (9); \draw [-Stealth] (2) -- (10);
		\draw [-Stealth] (3) -- (6); \draw [-Stealth] (3) -- (7); \draw [-Stealth] (3) -- (8); \draw [-Stealth] (3) -- (9); \draw [-Stealth] (3) -- (10);
		\draw [-Stealth] (4) -- (6); \draw [-Stealth] (4) -- (7); \draw [-Stealth] (4) -- (8); \draw [-Stealth] (4) -- (9); \draw [-Stealth] (4) -- (10);
		\draw [-Stealth] (5) -- (6); \draw [-Stealth] (5) -- (7); \draw [-Stealth] (5) -- (8); \draw [-Stealth] (5) -- (9); \draw [-Stealth] (5) -- (10);
		\draw [-Stealth] (6) -- (11); \draw [-Stealth] (6) -- (12);
		\draw [-Stealth] (7) -- (11); \draw [-Stealth] (7) -- (12);
		\draw [-Stealth] (8) -- (11); \draw [-Stealth] (8) -- (12);
		\draw [-Stealth] (9) -- (11); \draw [-Stealth] (9) -- (12);
		\draw [-Stealth] (10) -- (11); \draw [-Stealth] (10) -- (12);
	\end{tikzpicture}
	\hspace{5mm}
	\begin{tikzpicture}[]
		\node[circle,minimum size=15,draw](1) at (0,0) {};
		\node[circle,minimum size=15,draw](2) at (0,0.75) {};
		\node[circle,minimum size=15,draw](3) at (0,1.5) {};
		\node[circle,minimum size=15,draw](4) at (0,2.25) {};
		\node[circle,minimum size=15,draw](5) at (0,3) {};
		\node[circle,minimum size=15,draw](6) at (1.5,0) {};
		\node[circle,minimum size=15,draw](7) at (1.5,0.75) {};
		\node[circle,minimum size=15,draw](8) at (1.5,1.5) {};
		\node[circle,minimum size=15,draw](9) at (1.5,2.25) {};
		\node[circle,minimum size=15,draw](10) at (1.5,3) {};
		\node[circle,minimum size=15,draw](11) at (3,1) {};
		\node[circle,minimum size=15,draw](12) at (3,2) {};
		\draw [-Stealth] (1) -- (7); \draw [-Stealth] (1) -- (10);
		\draw [-Stealth] (2) -- (6); \draw [-Stealth] (2) -- (7);
		\draw [-Stealth] (3) -- (8); \draw [-Stealth] (3) -- (9);
		\draw [-Stealth] (4) -- (6);
		\draw [-Stealth] (5) -- (6); \draw [-Stealth] (5) -- (7); \draw [-Stealth] (5) -- (8); \draw [-Stealth] (5) -- (9);
		\draw [-Stealth] (7) -- (11); \draw [-Stealth] (7) -- (12);
		\draw [-Stealth] (8) -- (11); \draw [-Stealth] (8) -- (12);
		\draw [-Stealth] (10) -- (12);
	\end{tikzpicture}
	\caption{An example of dropconnect. The right network has had weights dropped with probability 0.5.}
	\label{fig:dropconnect}
\end{figure}

One of the first proposed variations on standard dropout was \textit{dropconnect}, introduced in 2013 by \citet{wan2013dropconnect}. This method is a generalization of dropout where individual weights and biases rather than neuron outputs are set to zero with some probability. So, in training, the output of a network layer is given by:
\begin{equation}
	\y = f((\W\circ\M)\x),  \quad m_{ij} \sim \text{Bernoulli}(1-p),
\end{equation}
where terms are defined as in \ref{eq:standard}, but with a dropout mask matrix rather than a vector. Dropconnect is illustrated in Figure~\ref{fig:dropconnect}.

Dropconnect takes a different approach than standard dropout during test time. Rather than setting weights to their average value, the authors propose a Gaussian approximation of dropconnect at each neuron~\cite{wan2013dropconnect}. A sample is then taken from this Gaussian and passed to the neuron activation function. This makes dropconnect a stochastic method at test time as well as during training. The authors show that dropconnect can regularize some networks more effectively than standard dropout, at the expense of requiring larger dropout masks. Dropout methods that drop weights rather than neurons are also sometimes called weight dropout.

Another area of improvement over standard dropout that has been explored is speeding up training convergence when using dropout. \textit{Fast dropout}~\cite{wang2013fast}, also proposed in 2013, provides a faster way to do dropout-like regularization, by interpreting dropout methods from a Bayesian perspective. The authors show that the outputs of layers with dropout can be seen as sampling from an underlying distribution, which can be approximated by a Gaussian. This distribution can then either be sampled from directly or its parameters can be used to propagate information about the entire dropout ensemble. This technique can lead to faster training procedures than standard dropout, where only one element of the ensemble of possible networks is sampled at once. One reason for this is that in standard dropout, for a given training sample, the fraction of neurons trained on the sample is \(p\). To effectively use an entire training dataset to train all neurons requires passing each sample through the network multiple times. Fast dropout exposes all neurons to each training sample, which avoids this slowdown. Fast dropout can also be directly applied at test time, as opposed to the approximate averaging employed in standard dropout. 

Several proposed dropout methods seek to improve regularization or speed up convergence by making dropout adaptive, that is tuning dropout probabilities during training based on neuron weights or activations. A major example is \textit{Standout}~\cite{ba2013adaptive}, again proposed in 2013. This method overlays a binary belief network onto a neural network which controls the dropout properties of individual neurons. The authors intepret the belief network as tuning the architecture of the neural network. For each weight in the original neural network, Standout adds a corresponding weight parameter in the binary belief network. A layer's output during training is given by:
\begin{equation}
	y = f(\W\x) \circ \m, \quad m_i \sim \text{Bernoulli}(g(\W_s\x)),
\end{equation}
where terms are defined as in (\ref{eq:standard}), but with \(\W_s\) representing the belief network's weights for that layer and \(g(\cdot)\) representing the belief network's activation function.

While a separate learning algorithm can be applied to learn the belief network weights, in practice, the authors found that this resulted in the belief network weights becoming approximately equal to an affine function of the corresponding neural network weights~\cite{ba2013adaptive}. So, an effective approach to determine belief network weights is setting them as
\begin{equation}
	\W_s = \alpha\W+\beta
\end{equation}
at each training iteration for some constants $\alpha$ and $\beta$.
The output of each layer during testing is given by:
\begin{equation}
	y = f(\W\x) \circ g(\W_s\x).
\end{equation}

Another adaptive dropout method, inspired by a Bayesian understanding of dropout, is \textit{variational dropout}, as proposed by \citet{kingma2015variational} in 2015. The authors show that a variant of dropout that uses Gaussian multiplicative noise (proposed by \citet{srivastava2014dropout}) can be interpreted as a variational method given a particular prior over the network weights and a particular variational objective. They then derive an adaptive dropout scheme that can automatically determine an effective dropout probability for an entire network, or for individual layers or neurons. Variational dropout has also been applied to sparsify networks as a step in model compression. This application is described in section \ref{sec:compression}.

Other significant adaptive dropout methods include \textit{evolutionary dropout}~\cite{li2016improved}, which uses second-order statistics of neuron activations across a minibatch to set dropout probabilities, and \textit{concrete dropout}~\cite{gal2017concrete}, which applies a variational interpretation of dropout to set dropout probabilities in a principled way.

A simple modification to standard dropout that some researchers have explored is changing the dropout probability according to a schedule during training. \textit{Annealed dropout}~\cite{rennie2014annealed} is a method in which the dropout probability is gradually reduced during training, with the goal of taking advantage of a larger effective network during later training iterations. It uses a schedule of the form
\begin{equation}
	p_t = \max\left(0,1-\frac{t}{N}\right)p_0,
\end{equation}
where \(p_t\) is the dropout probability used at training iteration \(t\), \(p_0\) is the initial dropout probability, and \(N\) is a constant. On the other hand, \citet{morerio2017curriculum} argue that increasing dropout probability during training is preferable, reasoning that stronger regularization is needed more later in training to avoid overfitting. They propose \textit{curriculum dropout}, which uses a schedule of the form
\begin{equation}
	p_t = p_\infty(1-e^{-\gamma t}),
\end{equation}
where \(p_\infty\) is the upper limit that \(p_t\) approaches as \(t \rightarrow \infty\) and \(\gamma\) is a constant.

\section{Convolutional layers} \label{sec:cnn}

Convolutional neural network layers require different regularization methods than standard dropout in order to generalize well~\cite{tompson2015efficient,he2016deep}. This is because the pixels in feature maps produced by a convolutional layer are highly correlated, and so randomly dropping some has little effect~\cite{tompson2015efficient}. However, many promising alternative approaches to using dropout as a regularization method for training CNNs have been proposed. These include applying dropout to larger regions than individual neurons, applying dropout at different places in the network topology, and designing dropout methods for deep residual networks.

\textit{Batch normalization}~\cite{ioffe2015batch} is another regularization technique commonly used with convolutional neural networks. Some authors have found that using batch normalization reduces or eliminates the benefits of using dropout for regularization~\cite{ioffe2015batch,he2016deep}. However, recently proposed dropout methods have often shown that when using a dropout method adapted specifically for convolutional layers, better results are achieved compared to batch normalization alone~\cite{park2016analysis,devries2017improved,huang2016deep,singh2016learning,cai2019effective}.

One approach for achieving strong regularization given highly correlated activations is to drop larger regions than individual pixels. \textit{Spatial dropout}, proposed in 2015, takes this approach~\cite{tompson2015efficient}~\cite{tompson2015efficient}. When using spatial dropout, instead of dropping individual pixels, entire feature maps are dropped with probability \(p\). This prevents the network from using nearby pixels to recover information when dropout is applied. The authors showed that this method improved performance on object localization~\cite{tompson2015efficient}. \citet{park2016analysis} propose an improvement to this where the spatial location or the feature map with the highest activation is dropped with some probability \(p_{\text{off}}\). This encourages the model to rely on a wider range of information when making classification decisions. They also propose to increase the robustness of the network to dropped neurons by sampling the dropout probability itself from some probability distribution at each iteration. The authors suggest either a normal distribution \(p \sim \mathcal{N}(\mu,\sigma)\) or a uniform distribution \(p \sim U(a,b)\). Their results show that these two modifications can improve results on image classification datasets~\cite{park2016analysis}.

Another CNN dropout method that works by dropping out larger regions is \textit{cutout}~\cite{devries2017improved}, which applies a random square mask over a region of each input image. Unlike other common methods which apply dropout at the feature map level, this method directly applies to the input image. The main motivation behind cutout is removing visual features with high activation values in later layers of a CNN. However, the authors argue that this masking approach on input images has equivalent performance and is cheaper to conduct~\cite{devries2017improved}. 

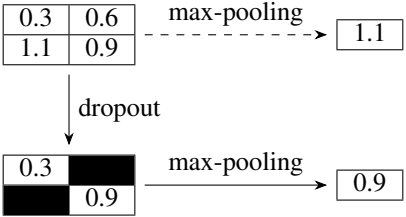
\begin{figure}[!t]
	\centering
	\begin{tikzpicture}[]
		\node[](a) at (0,2) {
			\begin{tabular}{|c|c|} \hline
				0.3 & 0.6 \\ \hline
				1.1 & 0.9 \\ \hline
			\end{tabular}
			};
		\node[](b) at (0,0) {
			\begin{tabular}{|c|c|} \hline
				0.3 & \cellcolor{black} \\ \hline
				\cellcolor{black} & 0.9 \\ \hline
			\end{tabular}
			};
		\node[](c) at (4,2) {\begin{tabular}{|c|} \hline 1.1 \\ \hline \end{tabular}};
		\node[](d) at (4,0) {\begin{tabular}{|c|} \hline 0.9 \\ \hline \end{tabular}};
		\draw[-Stealth] (a) -- (b) node[midway,right] {dropout};
		\draw[-Stealth,dashed] (a) -- (c) node[midway,above] {max-pooling};
		\draw[-Stealth] (b) -- (d) node[midway,above] {max-pooling};
	\end{tikzpicture}
		\caption{Max-pooling dropout in convolutional neural networks~\cite{wu2015towards}.}
	\label{fig:max_pooling}
\end{figure}

Another approach some authors have explored for improving dropout-based regularization in CNNs is to apply dropout at alternate points in the network topology. \textit{Max-pooling dropout}~\cite{wu2015towards} is one such method, where dropout is integrated into a max-pooling layer. Max-pooling is a common operation in CNN topologies which selects the maximum activation value from non-overlapping areas of an input feature map. This simplifies following layers at the cost of potentially losing useful information. Max-pooling dropout retains the behaviour of max-pooling layers while probabilistically allowing other feature values to affect the output of a pooling layer. This operator masks a subset of feature values before performing the max-pooling operation. As Figure~\ref{fig:max_pooling} shows, the max-pooling operator always pools the largest value in a given pooling window, while the max-pooling dropout method provides an opportunity for smaller feature values to affect activations in later layers. This technique can help the network to avoid overfitting as saturated activation values have less contribution in the network loss. At test time, the pooling operation becomes a linear sum over activations, where each activation is weighted by the probability that it would be selected as the output during training according to this dropout method.

\citet{cai2019effective} argue that the observed lack of additional regularization when using dropout along with batch normalization is simply due to incorrectly ordering those two operations. They examine using dropout alongside batch normalization in CNNs at neuron, channel, path, and layer levels. In their proposed method, dropout and batch normalization are reordered in convolutional building blocks to address the increase of variance from random deactivation of basic components such as neurons~\cite{cai2019effective}. The authors claim that the failure of standard dropout, resulting in training instability, is due to the incorrect placement of dropout and batch normalization operations in the the convolutional layer. The conducted experiments on various datasets show that reordering them improves performance~\cite{cai2019effective}.

The development of very deep convolutional neural networks using residual layers~\cite{he2016deep} has inspired new dropout methods for such networks. \textit{Stochastic depth} is a dropout method proposed by \citet{huang2016deep} where entire layers are dropped randomly during training, and values are instead passed through unchanged. This allows for extremely deep networks to be effectively trained. Another method based on residual networks is \textit{swapout}~\cite{singh2016learning}. This method operates on individual neurons by randomly selecting between the neuron output, the corresponding input, the sum of the input and output (a residual connection), and the value zero.

\section{Recurrent layers} \label{sec:rnn}


In general, feedforward dropout methods as described in section \ref{sec:training} can be applied to the feedforward connections of a network containing recurrent layers. Research has therefore focused on applying dropout methods to recurrent connections. Applying standard dropout to these connections results in poor performance~\cite{zaremba2015recurrent}, since the noise caused by dropout at each time step prevents the network from retaining long-term memory. However, methods that are specialized for recurrent layers have proved successful, and are commonly used in practice. Generally speaking, they apply dropout to recurrent connections in a way that can still preserve long-term memory.

Research into dropout in recurrent neural networks (RNNs) has focused on long short-term memory (LSTM) networks, although some proposed methods can be applied to RNNs in general. The following is a typical definition of an LSTM cell, although variations exist. For an input \(\x_t\) at time \(t\), input, forget, and output, gate signals are defined as:
\begin{align}
	\mathbf{i}_t &= \sigma\left(\W_i \x_t + \U_i \h_{t-1} \right),  \\  
	\mathbf{f}_t &= \sigma\left(\W_f \x_t + \U_f \h_{t-1} \right),
\end{align}
	and
\begin{equation}
	\mathbf{o}_t = \sigma\left(\W_o \x_t + \U_o \h_{t-1} \right),
	\end{equation}
	respectively. The cell state is defined as:
\begin{equation}
	\mathbf{c}_t = \mathbf{f}_t\circ\mathbf{c}_{t-1} + \mathbf{i}_t\circ\mathbf{g}_t,
	\end{equation}
	where 
\begin{equation}
	\mathbf{g}_t = \tanh\left(\W_g \x_t + \U_g \h_{t-1} \right).
	\end{equation}
The hidden state, which is the layer's output, is defined as:
\begin{equation}
	\h_t = \mathbf{o}_t\circ\tanh(\mathbf{c}_t).
	\end{equation}
\(\W\) and \(\U\) matrices represent learned weights, $\sigma(\cdot)$ represents a sigmoid activation function, and \(\sigma(\cdot)\) and \(\tanh(\cdot)\) are applied element-wise. For more information on LSTM networks, see \cite{salehinejad2017recent}.



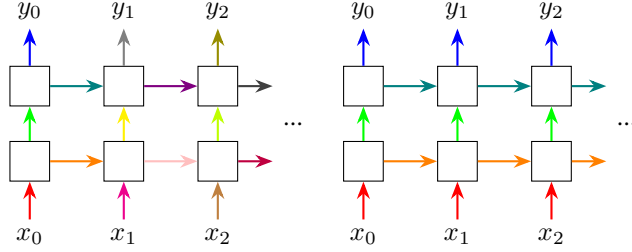
\begin{figure}[!t]
	\centering
	\begin{tikzpicture}[]
		\node[](x0) at (0,0) {\(x_0\)};
		\node[rectangle,minimum size = 15,draw](a0) at (0,1) {};
		\node[rectangle,minimum size = 15,draw](b0) at (0,2) {};
		\node[](y0) at (0,3) {\(y_0\)};
		\node[](x1) at (1.25,0) {\(x_1\)};
		\node[rectangle,minimum size = 15,draw](a1) at (1.25,1) {};
		\node[rectangle,minimum size = 15,draw](b1) at (1.25,2) {};
		\node[](y1) at (1.25,3) {\(y_1\)};
		\node[](x2) at (2.5,0) {\(x_2\)};
		\node[rectangle,minimum size = 15,draw](a2) at (2.5,1) {};
		\node[rectangle,minimum size = 15,draw](b2) at (2.5,2) {};
		\node[](y2) at (2.5,3) {\(y_2\)};
		\node[rectangle,minimum size = 15](a3) at (3.5,1) {};
		\node[rectangle,minimum size = 15](b3) at (3.5,2) {};
		\node[]() at (3.5,1.5) {...};
		\draw [red,-Stealth,thick] (x0) -- (a0); \draw [green,-Stealth,thick] (a0) -- (b0); \draw [blue,-Stealth,thick] (b0) -- (y0);
		\draw [magenta,-Stealth,thick] (x1) -- (a1); \draw [yellow,-Stealth,thick] (a1) -- (b1); \draw [gray,-Stealth,thick] (b1) -- (y1);
		\draw [brown,-Stealth,thick] (x2) -- (a2); \draw [lime,-Stealth,thick] (a2) -- (b2); \draw [olive,-Stealth,thick] (b2) -- (y2);
		\draw [orange,-Stealth,thick] (a0) -- (a1); \draw [pink,-Stealth,thick] (a1) -- (a2); \draw [purple,-Stealth,thick] (a2) -- (a3);
		\draw [teal,-Stealth,thick] (b0) -- (b1); \draw [violet,-Stealth,thick] (b1) -- (b2); \draw [darkgray,-Stealth,thick] (b2) -- (b3);
	\end{tikzpicture}
	\hspace{2mm}
	\begin{tikzpicture}[]
		\node[](x0) at (0,0) {\(x_0\)};
		\node[rectangle,minimum size = 15,draw](a0) at (0,1) {};
		\node[rectangle,minimum size = 15,draw](b0) at (0,2) {};
		\node[](y0) at (0,3) {\(y_0\)};
		\node[](x1) at (1.25,0) {\(x_1\)};
		\node[rectangle,minimum size = 15,draw](a1) at (1.25,1) {};
		\node[rectangle,minimum size = 15,draw](b1) at (1.25,2) {};
		\node[](y1) at (1.25,3) {\(y_1\)};
		\node[](x2) at (2.5,0) {\(x_2\)};
		\node[rectangle,minimum size = 15,draw](a2) at (2.5,1) {};
		\node[rectangle,minimum size = 15,draw](b2) at (2.5,2) {};
		\node[](y2) at (2.5,3) {\(y_2\)};
		\node[rectangle,minimum size = 15](a3) at (3.5,1) {};
		\node[rectangle,minimum size = 15](b3) at (3.5,2) {};
		\node[]() at (3.5,1.5) {...};
		\draw [red,-Stealth,thick] (x0) -- (a0); \draw [green,-Stealth,thick] (a0) -- (b0); \draw [blue,-Stealth,thick] (b0) -- (y0);
		\draw [red,-Stealth,thick] (x1) -- (a1); \draw [green,-Stealth,thick] (a1) -- (b1); \draw [blue,-Stealth,thick] (b1) -- (y1);
		\draw [red,-Stealth,thick] (x2) -- (a2); \draw [green,-Stealth,thick] (a2) -- (b2); \draw [blue,-Stealth,thick] (b2) -- (y2);
		\draw [orange,-Stealth,thick] (a0) -- (a1); \draw [orange,-Stealth,thick] (a1) -- (a2); \draw [orange,-Stealth,thick] (a2) -- (a3);
		\draw [teal,-Stealth,thick] (b0) -- (b1); \draw [teal,-Stealth,thick] (b1) -- (b2); \draw [teal,-Stealth,thick] (b2) -- (b3);
	\end{tikzpicture}
	\caption{Comparison of per-step (left) versus per-sequence (right) sampling of dropout masks on an unrolled RNN. Horizontal connections are recurrent while vertical connections are feedforward. Different colours represent different dropout masks applied to the corresponding connection.}
	\label{fig:rnn_step_vs_seq}
\end{figure}

\textit{RNNdrop}~\cite{moon2015rnndrop}, proposed in 2015, provides a simple solution to better preserve memory when applying dropout. The key change is to generate a dropout mask for each input sequence, and keep it the same at every time step. This varies from the naive way of applying dropout to RNNs, which would generate new dropout masks for each input sample, regardless of which time sequence it was from. Generating masks on a per-sequence basis means that the elements in the network hidden state that are not dropped will persist throughout the entire sequence without ever being affected by dropout, which allows the network to maintain long-term memory. The difference between per-step and per-sequence masks on an unrolled RNN is illustrated in Figure~\ref{fig:rnn_step_vs_seq}.

In particular, the authors propose applying dropout to the hidden cell state. So, the only change from the original LSTM definition is the equation for \(\mathbf{c}_t\), which becomes
\[
	\mathbf{c}_t = \m\circ(\mathbf{f}_t\circ\mathbf{c}_{t-1} + \mathbf{i}_t\circ\mathbf{g}_t), \quad m_{i} \sim \text{Bernoulli}(1-p).
\]

Various other proposed methods also use per-sequence dropout mask sampling on recurrent connections to help preserve long-term memory. \textit{Variational RNN dropout}~\cite{gal2016rnn}, proposed in 2016, is one such method, but it operates in a way that is theoretically justified in terms of a Bayesian interpretation of RNN dropout. The authors show that if dropout is seen as a variational Monte Carlo approximation to a Bayesian posterior, then the natural way to apply it to recurrent layers is to generate a dropout mask that zeroes out both feedforward and recurrent connections for each training sequence, but to keep the same mask for each time step in the sequence. This is similar to RNNdrop in that masks are generated on a per-sequence basis, but the derivation leads to dropout being applied at a different point in the LSTM cell. Formally, the equations for \(\mathbf{i}_t\), \(\mathbf{f}_t\), \(\mathbf{o}_t\), and \(\mathbf{g}_t\) become:
\begin{align}
	\mathbf{i}_t &= \sigma\left(\W_i(\x_t\circ\m_x) + \U_i(\h_{t-1}\circ\m_h)\right) \\
	\mathbf{f}_t &= \sigma\left(\W_f(\x_t\circ\m_x) + \U_f(\h_{t-1}\circ\m_h)\right) \\
	\mathbf{o}_t &= \sigma\left(\W_o(\x_t\circ\m_x) + \U_o(\h_{t-1}\circ\m_h)\right) \\
	\mathbf{g}_t &= \tanh\left(\W_g(\x_t\circ\m_x) + \U_g(\h_{t-1}\circ\m_h)\right) \\
	m_{x,i}&,m_{h,i} \sim \text{Bernoulli}(1-p)
\end{align}
with the equations for \(\mathbf{c}_t\) and \(\mathbf{h}\) remaining the same as in the original LSTM. This dropout method has become one of the most widespread techniques for regularizing RNNs.

One other proposed method using per-sequence mask sampling is \textit{weight-dropped LSTMs}, proposed in 2017~\cite{merity2017regularizing}. This method takes inspiration from dropconnect, rather than standard dropout, also dropping out weights rather than activations. In training, these LSTM cells use the following equations for \(\mathbf{i}_t\), \(\mathbf{f}_t\), \(\mathbf{o}_t\), and \(\mathbf{g}_t\), otherwise following the basic LSTM formulation given above.
\begin{align}
	\mathbf{i}_t &= \sigma\left(\W_i\x_t + (\U_i\circ\M)\h_{t-1}\right) \\
	\mathbf{f}_t &= \sigma\left(\W_f\x_t + (\U_f\circ\M)\h_{t-1}\right) \\
	\mathbf{o}_t &= \sigma\left(\W_o\x_t + (\U_o\circ\M)\h_{t-1}\right) \\
	\mathbf{g}_t &= \tanh\left(\W_g\x_t + (\U_g\circ\M)\h_{t-1}\right) \\
	M_{ij} &\sim \text{Bernoulli}(1-p)
\end{align}
This approached allows the authors to achieve results on language modelling benchmarks that were state-of-the-art at the time~\cite{merity2017regularizing}.


\textit{Recurrent dropout}~\cite{semeniuta2016recurrent} is an alternative approach that can preserve memory in an LSTM while still generating different dropout masks for each input sample, as in standard dropout. This is done by only applying dropout to the part of the RNN that updates the hidden state and not the state itself. So, if an element is dropped, then it simply does not contribute to network memory, rather than erasing the hidden state. For an LSTM, the equations are the same as in the original LSTM except that the equation for \(\mathbf{c}_t\) becomes
\begin{equation}
	\mathbf{c}_t = \mathbf{f}_t\circ\mathbf{c}_{t-1} + \mathbf{i}_t\circ\mathbf{g}_t\circ\m_t, \quad m_{t,i} \sim \text{Bernoulli}(1-p).
\end{equation}

Another proposed dropout method that can help to preserve memory in RNNs is \textit{Zoneout}~\cite{krueger2016zoneout}. This method randomly replaces neuron activations with the corresponding activations from the previous time-step. The authors interpret this method as being related to stochastic depth and swapout, as discussed in Section \ref{sec:cnn}, but with information being stochastically passed through from previous timesteps as opposed to previous layers.

\section{Dropout methods for model compression} \label{sec:compression}

Standard dropout promotes sparsity in neural network weights~\cite{srivastava2014dropout}. This property means that dropout methods can be applied in compressing neural network models by reducing the number of parameters needed to perform effectively. Since 2017, several dropout-based approaches have been proposed for practical model compression.

In 2017, \citet{molchanov2017variational} proposed using variational dropout~\cite{kingma2015variational} (described in Section~\ref{sec:training}) to sparsify both fully connected and convolutional layers. This approach was shown to achieve large reductions in the number of parameters in standard convolutional networks while minimally affecting performance. This sparse representation can then be passed into existing methods that convert sparse networks into compressed models, as in \cite{han2016deep}. A similar method was proposed by \citet{neklyudov2017structured}, which uses a modified variational dropout scheme that promotes sparsity, but the resulting network is specifically structured in such a way that is easy to compress.


Developing further dropout methods for model compression has been an area of significant activity recently. Recently proposed approaches include \textit{targeted dropout}~\cite{gomez2018targeted}, in which neurons are chosen adaptively to be dropped out in such a way that the network adapts to neural pruning, allowing it to be shrunk considerably without much loss in accuracy. Another recent proposal is \textit{Ising-dropout}~\cite{salehinejad2019ising}, which overlays a graphical Ising model on top of a neural network in order to identify less useful neurons, and drop them out in both training and inference. We expect to continue to see advances in applying dropout methods for model compression.

\section{Monte Carlo dropout} \label{sec:mc}

In many machine learning tasks, it is useful to know how certain a model's output is. For instance, a classification output is more likely to be correct when an input is very similar to elements of the training set than when its input is dissimilar to all training data. Most neural network models do not provide this information. Bayesian machine learning models, on the other hand, often produce outputs that are probability distributions, giving more information about model certainty~\cite{gal2016uncertainty}. \textit{Monte Carlo dropout} is a dropout method that can produce model uncertainty estimates in an analogous way~\cite{gal2016dropout}.

In 2016, \citet{gal2016dropout} proposed a Bayesian theoretical understanding of dropout that has since become widely accepted. They interpret dropout as a sampling method that is equivalent to a variational approximation of a deep Gaussian process. A deep Gaussian process is a Bayesian machine learning model that would normally produce a probability distribution as its output, and applying standard dropout at test time (rather than scaling weights and using all neurons as described in Section~\ref{sec:standard}) can be used to estimate characteristics of this underlying distribution. The estimated variance of the distribution is taken to indicate the uncertainty of the model for a particular input. This method of estimating uncertainty is called Monte Carlo dropout.

To implement Monte Carlo dropout, a neural network is first trained normally using standard dropout. To perform inference on an input sample, the network is run \(T\) times with standard dropout, all with the same input but with different randomly generated dropout masks each time. Estimators for the mean and variance of the implicit Bayesian model output are given by~\cite{gal2016uncertainty}:
\begin{align*}
	\E[\y] &\approx \frac{1}{T} \sum_{t=1}^T \hat{\y}_t(\x) \\
	\Var(\y) &\approx \tau^{-1}\ident_D + \frac{1}{T} \sum_{t=1}^T \hat{\y}_t(\x)^T\hat{\y}_t(\x) - \E[\y]^T\E[\y],
\end{align*}
where \(\hat{\y}_t(\x)\) is the output of the network given inputs \(\x\) and the \(t\)th set of dropout masks and \(\tau\) is a constant determined by the model structure. These are respectively taken to be the model output and an indication of the model uncertainty.

Monte Carlo dropout has found many applications in practice, including in time-series prediction~\cite{zhu2017deep} and medical imaging~\cite{jungo2018towards}. Other approaches proposed to measure model uncertainty include Bayesian neural networks~\cite{gal2016uncertainty} and ensemble-based approaches\cite{lakshminarayanan2017simple}. Monte Carlo dropout has an advantage over these methods in that changes are not needed to the model training procedure, whereas both of these approaches incur a large increase in training complexity.

\section{Discussion} \label{sec:discussion}

We have described a wide range of advances in dropout methods above. This section discusses ongoing research trends in broader terms.

The most common research direction in dropout methods has been improving dropout for regularization. It is generally accepted that standard dropout can regularize a wide range of neural network models, but there is room to achieve either faster training convergence or better final performance. The former concern is important since dropout reduces the exposure of neurons to each training sample, which can slow down training~\cite{wang2013fast}. With neural networks becoming larger and more computationally intensive to train, techniques such as fast dropout~\cite{wang2013fast} that reduce this effect are valuable. Improving how dropout affects the performance of trained networks is also an ongoing concern. Trying to drop neurons in a more intelligent or theoretically justified way than standard dropout has shown promise. Also, the growth of convolutional and recurrent neural networks in practice has prompted the development of specialized methods that perform better than standard dropout on specific kinds of neural networks. As new kinds of neural networks and neural network layers continue to be developed, there continue to be opportunities to design or improve on specialized dropout methods.

Other research into dropout methods looks to widen their applications beyond regularization. As discussed above, this includes the use of dropout for model compression, either on its own or in concert with existing model compression techniques. As with regularization, there are opportunities to develop improved methods that are specialized for particular kinds of networks or that use more advanced approaches for selecting neurons to drop. Monte Carlo dropout is another application of dropout methods: using them to measure model uncertainty. There is potential for further applications, given the broad ability of dropout methods to stochastically guide network training and operation.

A promising line of research has emerged into adversarial dropout methods~\cite{park2018adversarial,saito2018adversarial,park2019adversarial}. These techniques either incorporate dropout methods into adversarial learning schemes, or apply ideas from adversarial learning to guide dropout procedures in a more effective way.

Finally, a substantial amount of theoretical analysis has been done to rigorously justify existing dropout methods. The growth of Bayesian interpretations of dropout methods over the last few years points to new opportunities in theoretical justifications of dropout and similar stochastic methods, which corresponds to a broader trend of Bayesian and variational techniques advancing research into deep neural networks.

In general, dropout methods have continually shown their utility and potential throughout deep learning, and we expect this trend to continue as deep neural networks continue to become more advanced and widely used.

\nocite{wardefarley2014empirical,baldi2013understanding,goodfellow2013maxout,rennie2014annealed,li2017dropout,morerio2017curriculum,li2016improved,saito2018adversarial,park2018adversarial,park2019adversarial,zolna2018fraternal,achille2018information,gomez2018targeted,khan2019regularization,hou2019weighted}

\bibliographystyle{IEEEtranN}
\bibliography{mybibfile}

\end{document}